\pdfoutput=1

\documentclass[11pt]{article}

\usepackage{acl}
\usepackage[ruled,vlined]{algorithm2e}
\usepackage{multirow}
\usepackage{times}
\usepackage{latexsym}
\usepackage[acronym]{glossaries}
\usepackage{graphicx}
\usepackage{adjustbox}
\usepackage[T1]{fontenc}

\usepackage[utf8]{inputenc}
\usepackage[ruled,vlined]{algorithm2e}
\SetKwProg{Fn}{Function}{:}{}
\SetKw{KwRet}{return}
\usepackage{microtype}
\usepackage{stfloats}
\title{Computational Analysis of Semantic Connections Between Herman Melville’s Reading and Writing}

\author{
Nudrat Habib, Elisa Barney Smith\\
Luleå University of Technology \\
\texttt{nudrat.habib@ltu.se} \\
\texttt{elisa.barney@ltu.se} 
\And
Steven Olsen-Smith \\
Boise State University \\
\texttt{sosmith@boisestate.edu}
}

\makeglossaries

\newacronym{ml}{ML}{Machine Learning}
\newacronym{llm}{LLM}{Large Language Model}
\newacronym{sota}{SotA}{state-of-the-art}
\newacronym{iaa}{IAA}{inter-annotator agreement}
\newacronym{docvqa}{DocVQA}{Document Visual Question Answering}
\newacronym{lv}{LV}{Language-Vision}
\newacronym{dia}{DIA}{Document Image Analysis}
\newacronym{vqa}{VQA}{Visual Question Answering}
\newacronym{idvqa}{iDocVQA}{Instruction Document Visual Question Answering}
\newacronym{pali}{PaLI}{Pathways Language and Image}
\newacronym{clip}{CLIP}{Contrastive Language–Image Pre-training}
\newacronym{llava}{LLaVA}{Large Language and Vision Assistant}
\newacronym{llama}{LLaMA}{Large Language Model Meta AI}
\newacronym{lmm}{LMM}{Large Multimodal Model}
\newacronym{mlp}{MLP}{multilayer perceptron}
\newacronym{beit3}{BEiT3}{BERT Pre-Training of Image Transformers-3}
\newacronym{nlp}{NLP}{natural language processing}
\newacronym{cv}{CV}{computer vision}
\newacronym{donut}{Donut}{Document Understanding Transformer}
\newacronym{cnn}{CNN}{Convolutional Neural Network}
\newacronym{ocr}{OCR}{optical character recognition}
\newacronym{ie}{IE}{information extraction}
\newacronym{ir}{IR}{information retrieval}
\newacronym{cord}{CORD}{consolidated receipt dataset}
\newacronym{llad}{LLaDoc}{Large Language Document}
\newacronym{squad}{SQuAD}{Stanford Question Answering Dataset}
\newacronym{spdvqa}{SP-DocVQA}{Single Page Document Visual Question Answering}
\newacronym{tvqa}{TextVQA}{Text Visual Question Answering}
\newacronym{pope}{POPE}{Polling-based Object Probing Evaluation}
\newacronym{CoT}{COT}{Chain-of-Thought}
\newacronym{LMs}{LM}{Language models}
\newacronym{wandb}{WandB}{weights and biases}
\newacronym{mpc}{MSRP}{Microsoft Research Paraphrase Corpus}
\newacronym{svm}{SVM}{Support Vector Machine}
\newacronym{tfidf}{TF-IDF}{Term Frequency-Inverse Document Frequency}
\newacronym{mt}{MT}{Machine Translation}

\begin{document}
\maketitle
\begin{abstract}
This study investigates the potential influence of Herman Melville’s reading on his own writings through computational semantic similarity analysis. Using documented records of books known to have been owned or read by Melville, we compare selected passages from his works with texts from his library. The methodology involves segmenting texts at both sentence level and non-overlapping 5-gram level, followed by similarity computation using BERTScore. Rather than applying fixed thresholds to determine reuse, we interpret precision, recall, and F1 scores as indicators of possible semantic alignment that may suggest literary influence. Experimental results demonstrate that the approach successfully captures expert-identified instances of similarity and highlights additional passages warranting further qualitative examination. The findings suggest that semantic similarity methods provide a useful computational framework for supporting source and influence studies in literary scholarship.
\end{abstract}

\section{Introduction}


Similarities between literary texts can arise for many different reasons. In some cases they may reflect the influence of earlier authors on later ones, while in others they result from shared ideas, stylistic conventions, or overlapping cultural contexts. Identifying such relationships is a necessary undertaking in literary and computational studies because it helps reveal how authors engage, consciously or unconsciously, with existing works. When dealing with historical documents, however, the sheer quantity of potentially relevant textual data presents serious obstacles to our comprehension of what materials an author might have encountered during their lifetime. Computational methods make it possible to trace potential textual relationships across large digital corpora. 

Of the three most frequently downloaded authors on Project Gutenberg, this study focuses on Herman Melville, a prominent 19\textsuperscript{th} century
 American author whose materials are widely accessible in digital form. Our investigation examines both the books Melville authored and those he is known to have read. Melville typically autographed his books when he acquired them, often inscribing the date of acquisition along with the city or immediate setting where he recorded the autograph inscription \cite{olsen2022books}. From the books of his collection that survive today, Melville’s inscribed date of acquisition supplies a chronological baseline for dating his reading of books that may have influenced works he went on to write. Even in cases of books from his library that are not known to survive, Melville’s practice of referring to book purchases in dated letters and journals can furnish similar evidence of acquisition \cite{olsen2022books}. When no external or holograph evidence exists to document Melville’s ownership or consultation of a potentially influential book, scholars are tasked to demonstrate its influence through methods of text comparison to infer possible relationships between his reading and his writing \cite{Bercaw1987}. Melville’s characteristic methods of source appropriation are especially well suited for the study of literary influence in terms both of detectability and of investigative outcome. As Harrison Hayford remarked in his study of Melville’s second book Omoo, the passages he derived from sources were “always rewritten and nearly always improved” by the process of appropriation \cite{Hayford1969Omoo}. Yet his methods of borrowing and altering from other books frequently left intact phrasal constructions and wording that make it possible to trace derived passages back to their original sources. This is of particular value in cases where no manuscripts of Melville’s works or source books marked and drawn upon by him are known to survive. The combination in Melville’s writings of vestigial source text and his creative reworking of that material illuminate his processes of literary composition as well as his ideological departures from his printed sources, exposing depths of authorial intention and varieties of artistic craft that can only be spotted through these investigations \cite{norberg2023technical}.
 
Traditional methods of identifying source influence involved a scholar's close reading and comparison of a created work and source books suspected to have played a role in its composition. While the scholar cannot be replaced, this process can be facilitated by modern tools. We aim to utilize computational tools to identify and analyze sources of textual influence in Melville's works and present them to scholars for further analysis.
Computational tools for evaluating textual similarity exist and are used for tasks like author attribution and plagiarism detection.  
We focus on semantic similarity analysis as a means of identifying potential traces of literary influence between Melville's works and books he owned or consulted in the process of composition. By framing the task in terms of automatic text comparison with reference sources, we draw on established computational techniques while shifting the emphasis from plagiarism detection, something needed in a modern context, to the scholarly question of how an author's use of source texts can be systematically discovered and evaluated.  By considering different forms of textual similarity ranging from word-for-word overlap to paraphrastic resemblances we argue that semantic similarity measures provide a flexible and powerful approach, particularly when the exact nature of influence is unknown in advance.



The research question we address is.

\textit{Can semantic similarity analysis identify traces of literary influence in Herman Melville’s writings when compared to books he owned or may otherwise have consulted in the course of composition?} 

Our contributions are as follows
\begin{itemize}
    \item We position semantic similarity analysis, and in particular BERTScore, as a  tool for investigating  the possible influence on an author's writing  by works  he read. 
    \item We present a feasibility study comparing selected passages from Melville’s works with texts from his documented readings, testing whether semantic similarity measures can highlight instances of overlap.
    \item Our findings provide insight into the strengths and limitations of using BERTScore in this context, laying the groundwork for scaling up to full corpus analyses and comparing our approach with alternative similarity metrics.
\end{itemize}


The rest of this paper  is organized as follows.
Section 2 reviews related work on textual similarity, intertextual analysis, and computational approaches to literary influence.
Section 3 describes the methodology, including data selection, segmentation strategies, and the use of BERTScore for semantic similarity analysis.
Section 4, titled Semantic Similarity Results: Evidence of Possible Influence, presents the experimental results and discussion.
Section 5 outlines the limitations of the current study and outlines the directions for future research, and
Section 6 concludes this research.

\section{Related Work}
Text similarity measures how closely two pieces of text match each other. In its extreme it can identify cases of plagiarism and many tools have been developed for this task. A high degree of textual similarity   does not mean plagiarism. It can reflect overlapping language or common expressions that require careful human interpretation to assess intent and originality \cite{meo2019turnitin}.






N-gram analysis has been a fundamental tool in text comparison. 
A study related to the performance evaluation on number of tokens used in an n-gram was done by \cite{clough2011developing}. 
\cite{clough2011developing} acknowledged that as the number n in an n-gram increases, the probability of finding commonalities in the documents decreases. 
They observed the worst results for n-grams above 5. 
They also recommended that the reference document be split into similar n-grams in the suspicious document. 
In n-gram comparison, possible plagiarism is recorded when a given threshold is surpassed \cite{clough2011developing}. 
\cite{shrestha2013using} further compared variants that removed or retained stop words and explored the use of named-entity n-grams, while  \cite{altheneyan2020automatic} highlighted the effect of punctuation and sentence-boundary errors on classification accuracy. These studies emphasize how segmentation and pre-processing choices influence detection outcomes.

Beyond surface matching, researchers have explored statistical and machine-learning approaches to text comparison.
\cite{vsaric2012takelab} employed correlation-based metrics such as overall Pearson and weighted mean for sentence-level similarity scoring. \cite{clough2011developing} applied Naive Bayes classification and evaluated performance using the F1 score, whereas  \cite{altheneyan2020automatic} implemented Support Vector Machines (SVMs) with lexical, syntactic, and semantic features. The Meteor metric, originating in machine translation, has also been adapted to detect paraphrastic or semantically equivalent passages \cite{altheneyan2020automatic}.
Although these studies were primarily designed for plagiarism detection, the underlying computational methods are equally valuable for investigating literary source use. 

Source and influence studies have a long tradition in pre-computational literary scholarship, starting in earnest roughly a century ago with John Livingston Lowe’s The Road to Xanadu, a study of Samuel Taylor Coleridge’s use of sources in his poetry \cite{lowes1927road}. Traditional methods of comparison continue to be carried out—often with significant results. Yet no work to date has systematically applied machine reading or automated analysis to examine how authors draw upon their sources in composing their own works. Broad digital applications to reading practices—most notably the Archaeology of Reading (which transcribes and exposes marginalia in machine-readable form) and Annotated Books Online \cite{annotated_books, archaeology_reading}, have made authorial reading methods and explicit annotations accessible for querying and close analysis. Simultaneously, single-author initiatives such as The Walt Whitman Archive, The William Blake Archive and Melville’s Marginalia Online have built extensive image and transcription corpora that encode the particular books an author actually handled and marked, steadily building the provenance and reading evidence necessary for undertakings like the present study \cite{whitman_archive, blake_archive, melville2006melville}. As these large-scale efforts finally achieve critical masses of machine-readable data, our research is now in a position to demonstrate the potential of computational approaches, providing new insights about source-use that are both textually grounded and empirically testable. These methods allow researchers not only to assess and verify instances where Melville and other major authors demonstrably engaged with known sources but also to identify previously unrecognized influences that may have shaped his writing.

The BERTScore was introduced by \cite{zhang2019bertscore} as an automatic evaluation metric for text generation. It measures semantic similarity between two texts by comparing their contextualized token embeddings produced by pretrained transformer models.  BERTScore can capture paraphrastic and meaning-level correspondence that extends beyond surface lexical overlap. In their original work, \cite{zhang2019bertscore} demonstrated that BERTScore correlates more strongly with human judgments than existing metrics (BLEU and METEOR) and provides improved model selection performance for tasks such as machine translation and image captioning.
Building on this foundation, \cite{tanahashi2020improving} applied BERTScore to the evaluation of translations of Japanese novels, further highlighting its usefulness for assessing semantic fidelity in literary contexts. More recently, \cite{jaskowski2024bertscorevisualizer} noted that current implementations of BERTScore do not expose all of the information the metric can produce, and introduced BERTScoreVisualizer, a tool that provides detailed token-level visualizations beyond the standard precision, recall, and F1 outputs. Together, these studies underscore the importance of BERTScore for semantic similarity analysis and highlight its effectiveness across a range of text comparison tasks. While most prior work has applied BERTScore in machine translation and related generation settings, its ability to quantify semantic similarity makes it a strong candidate for investigating possible literary influence between an author’s reading and writing.

While prior work has demonstrated the effectiveness of embedding-based metrics for measuring semantic similarity, their application to literary source and influence studies remains limited. This study builds on these advances by applying semantic similarity analysis to examine potential textual relationships between an author’s reading and writing.

\section{Methodology}
The broader goal of this research is to investigate semantic similarity across all works authored by Herman Melville and to compare them with the hundreds of books he is known to have read or owned in his personal library, as documented in \cite{Bercaw1987, Sealts1988}.

Conducting a large-scale similarity analysis across Melville’s complete corpus and all of his known library holdings presents significant computational and methodological challenges. Each of these works span hundreds of pages, and the number of possible pairwise comparisons is prohibitively large.

As a preliminary step, this study focuses on evaluating whether the proposed similarity analysis method can effectively detect meaningful textual overlaps. Specifically, we aim to determine whether the approach can identify known instances of semantic correspondence between Melville’s writings and the texts he read.

To achieve this, we first compiled a set of example text pair instances in which potential similarities were believed to exist based on scholarship by experts analyzing source appropriation \cite{Mathiesser1941AmericanRenaissance, olsen2011melville, bezanson1982historicalnote}. From these we selected four text instances.
The books and the corresponding text segments used in our experiments are detailed in Table \ref{instances}. 
These identified correspondences served as our ground truth for evaluation.

To ensure robustness, we did not restrict our analysis to only the passages whose correspondence has previously been reported. Instead, we extended the context by including text segments occurring both before and after each candidate and target passage. This approach allowed for a more realistic test of the similarity model’s capacity to recognize relevant relationships in a broader and less constrained textual environment. The structure of the books varies widely, with some short chapters and some long and some have no formal divisions at all. We thus did not extract uniform text lengths or sections across all text pair instances. Instead, we ensured that each extracted passage included sufficient context to allow for a meaningful similarity comparison, whether or not actual similarity was present. 

The texts for both Melville’s writings and his library holdings were obtained from Project Gutenberg\footnote{gutenberg.org} and the Internet Archive\footnote{archive.org}.
After identifying candidate segments, we then divided them into smaller units suitable for computational comparison. Two different chunking strategies were applied:
\begin{itemize}
    \item \textbf{N-gram based segmentation:}  We use the same n-gram size for both texts; the segments from Melville’s writings and the corresponding segments from the books he read, following the approach of \cite{shrestha2013using}. This method captures fine-grained lexical overlap and allows us to detect shorter recurring phrases or patterns that confirm borrowing from sources identified in the scholarship. 
    We choose the n-gram of 5 (5-gram), similar to the uses in \cite{suchomel2012three}. 
    We use non-overlapping n-grams. This has the advantage of less computation effort with a trade-off on accuracy as compared to overlapping window \cite{shrestha2013using}. We do not focus on named entity n-grams because they may not be common in the documents, thereby contributing less to detections \cite{shrestha2013using}. 
    \item \textbf{Sentence level segmentation:}  The selected segments were also divided at the sentence boundary. This provides larger semantic units that preserve syntactic and contextual information but will result in uneven chunk lengths. 
\end{itemize}
Using both chunking methods allowed us to examine similarity at different granularities, short phrase-level and longer, sentence-level.

\begin{table*}[t]
\footnotesize

\renewcommand{\arraystretch}{1.3}
\caption{Reference books and their corresponding mentions in Melville's works with target pages/chapters used in the comparison study.}
\label{instances}
\centering
\begin{adjustbox}{max width=\textwidth}
\begin{tabular}{|p{0.5cm}|p{4.6cm}|p{2cm}|p{2.2cm}|p{4.6cm}|p{2.2cm}|}
\hline
\textbf{S.No} &  \textbf{Reference Book} & \textbf{Author} & \textbf{Pages / Chapters} & \textbf{Melville's Book} & \textbf{Pages / Chapters} \\
\hline
1 & \textit{A Visit to the South Seas: In the U. States Ship Vincennes, During the Years 1829 and 1830, Volume 1 } & C.S. Stewart & Letter XXXIII 

(pp. 306--318) & \textit{Typee; or, Narrative of a Four Months' Residence } & Chapter XXV \\
\hline
2 & \textit{Life and Remarkable Adventures of Israel R. Potter } & Henry Trumbull & pp. 30--45 & \textit{Israel Potter; His Fifty Years of Exile } & Chapter 4 \\
\hline
3 & \textit{Religio Medici} & Sir Thomas Browne & 3 chapters  & \textit{Mardi; and a Voyage Thither } & 3 chapters  \\
\hline
4 & \textit{The Natural History of Sperm Whale} & Thomas Beale & Introductory 

remarks & \textit{Moby Dick} & Chapter 32 \\
\hline
\end{tabular}
\end{adjustbox}
\end{table*}

\begin{algorithm}[htbp]
\caption{Algorithm for Similarity Analysis between what Melville read and what he wrote}
\small                         
\DontPrintSemicolon
\SetAlgoLined
\setlength{\algomargin}{0.5em} 
\SetInd{0.5em}{0.8em}          

\KwIn{Folders:                                                 
\\ \quad \quad $M=\{M_1..M_n\}$ (Melville) 
\\ \quad \quad $R=\{R_1..R_n\}$ (Read)    
\\ \quad \quad $S=\{\text{5G},  \text{sent}\}$ Modes:  
\\ \quad \quad \quad 5G = non-overlapping 5-grams; 
\\ \quad \quad \quad sent = sentences}
\KwOut{Scored CSVs for each mode}

\textbf{Text segmentation (SEG)}
\SetKwFunction{SEG}{SEG}
\SetKwFunction{BERTSCORE}{BERTSCORE}

\Fn{\SEG{file, mode}}{
  \KwRet segments by mode \;
}

\ForEach{mode $\in S$}{
  \For{$i \leftarrow 1$ \KwTo $n$}{
    $X \leftarrow$ \SEG($M_i$, mode)\;   
    $Y \leftarrow$ \SEG($R_i$, mode)\;   
    $N \leftarrow \max(|X|, |Y|)$\;
    create CSV with header [ID, text, read\_text]\;
    \For{$k \leftarrow 1$ \KwTo $N$}{
      write row [$k$, $X_k$ or "", $Y_k$ or ""]\;
    }
    save CSV as merged\_i\_mode\;
  }

    \textbf{Computing Precision(p), Recall(r) and F Score(f) using BERTSCORE}

  \ForEach{CSV $\in$ \{merged\_1\_mode..merged\_n\_mode\}}{
    $T \leftarrow$ CSV[text]; \ $U \leftarrow$ CSV[read\_text]\;
    \For{$i \leftarrow 1$ \KwTo $|T|$}{           
      \For{$j \leftarrow 1$ \KwTo $|U|$}{         
        $(p,r,f) \leftarrow$ \BERTSCORE($T_i$, $U_j$)\;
        store $(p,r,f)$ for iteration $i$\;
      }
      append mean$(p,r,f)$ as columns $p_i, r_i, f_i$ in CSV\;
    }
    save CSV as scored\_i\_mode\;
  }
}
\label{Algorithm}
\end{algorithm}

For similarity analysis we used the BERTScore, a metric that evaluates the semantic similarity between two pieces of text using contextualized embeddings from pretrained transformer models \cite{zhang2019bertscore}. Unlike traditional lexical overlap metrics such as BLEU or ROUGE, BERTScore captures the meaning of words in context, making it particularly suitable for identifying subtle or paraphrased similarities across texts \cite{rostam2025evaluating}. This is especially important for our study, as  Melville's rewriting of source passages often produces alterations in vocabulary and phrasing, even when conveying similar ideas. BERTScore allows us to go beyond surface-level matching and evaluate whether two passages express similar meanings, regardless of exact duplication. Given that the influence of a text may appear through paraphrase, thematic similarity, or conceptual overlap rather than direct quotation, BERTScore is a well-suited choice for our task.

BERTScore produces precision, p, recall, r, and F1 scores (harmonic mean of p and r) that quantify the degree of semantic similarity between two texts. In prior work, particularly in evaluation of machine translation, higher scores are typically associated with closer semantic correspondence between a candidate text and a reference text. However, the literature does not establish any universal threshold that would indicate copying, reuse, or plagiarism, nor do such thresholds always apply straightforwardly to  studies of literary influence. In the context of this research, we therefore do not interpret BERTScore values as binary indicators of reuse. Instead, we treat relatively higher scores as signals of potential semantic alignment, which may suggest areas where thematic, phrasal, or conceptual influence could be present. Our aim is to identify passages where the model detects meaningful similarity that merits qualitative attention by literary scholars for possible literary influence.
We used the Microsoft/deberta-xlarge-mnli model variant of BERTScore, which has shown strong performance in natural language understanding tasks \cite{dernbach2024glam}. The algorithm of our methodology for text comparison is shown in Algorithm~\ref{Algorithm}. 

\section{Semantic Similarity Results: Evidence of Possible Influence}
Our experimental results demonstrate that the model reliably captured the  similarities in correspondences identified in the scholarship and also surfaced additional segments with high semantic similarity that had not been previously marked. In the subsections that follow, we examine these results across the different experimental settings and discuss expert identified instances at both n-gram and sentence level granularities, then model identified instances at both granularities. Finally we compare difference in results at both granularities.

\subsection{Expert Identified Instances: Validation of Semantic Similarity}
\begin{figure*}[h!]
\centering
\includegraphics[width=0.8\textwidth]{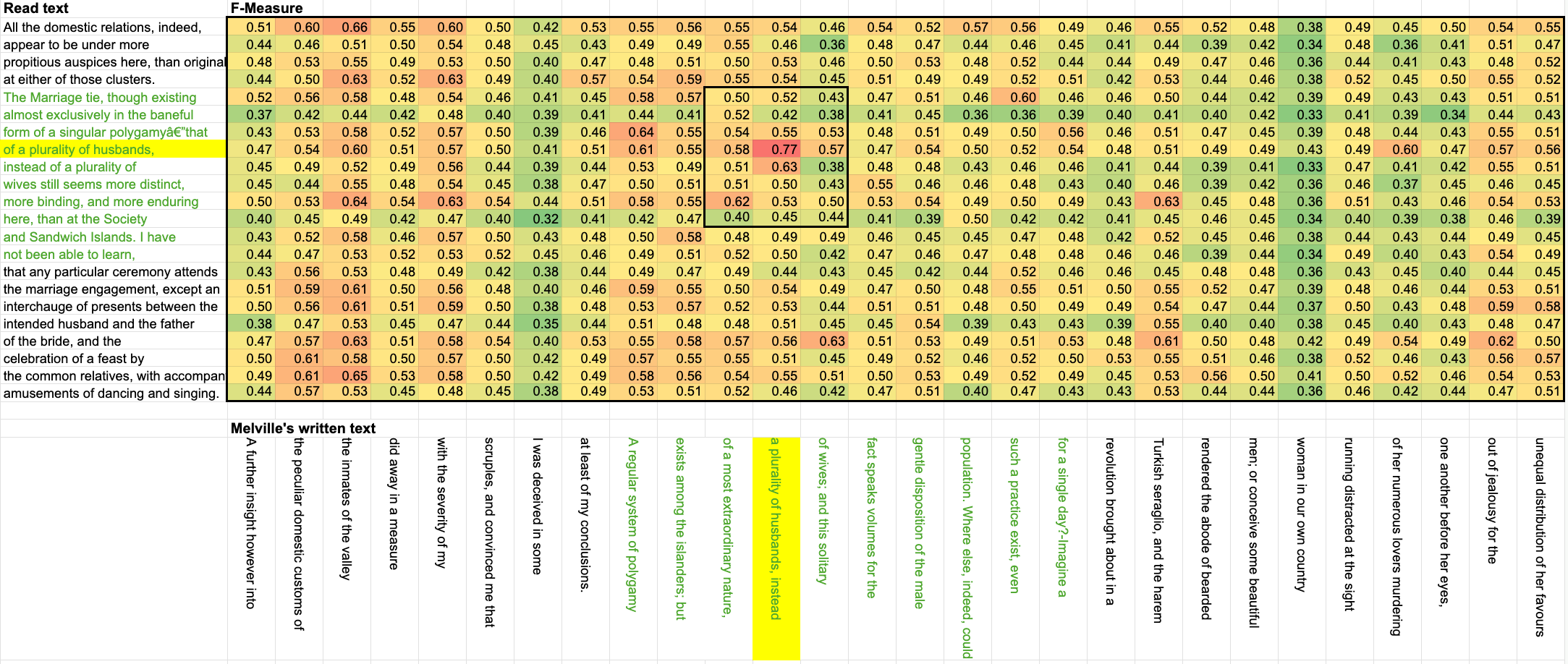}
\caption{5-gram analysis: F1 scores for instance 1 in Table \ref{instances} comparing \textit{Typee}  with  \textit{A visit to the south seas}. }
\label{polygamy_5}
\end{figure*}

\begin{figure*}[h!]
\centering
\includegraphics[width=0.8\textwidth]{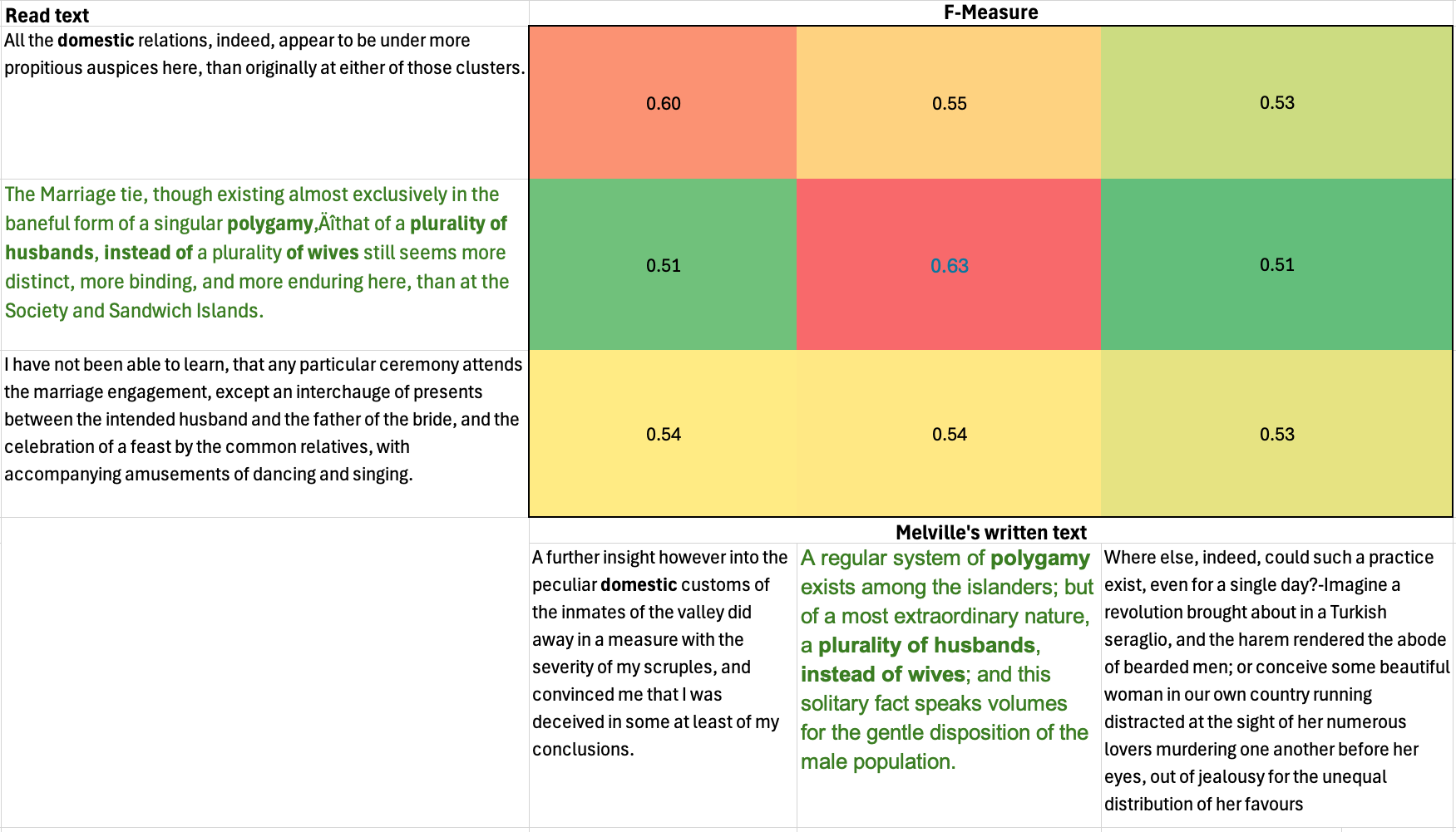}
\caption{Sentence level analysis: F1 scores for instance 1 in Table \ref{instances} comparing \textit{Typee}  with  \textit{A visit to the south seas}.}
\label{polygamy_sentences}
\end{figure*}

To illustrate how the model performs on an expert identified instance of potential influence, we consider the first instance in Table~\ref{instances}, where Melville's book \textit{Typee} is compared with Stewart's book \textit{A visit to the south seas}.
For illustrative purposes, throughout this paper only a selected portion of the experimental output from the text subsection is presented. The remaining results, comprising a broader set of sentence, and 5-gram-level comparisons, are omitted to maintain clarity and focus on representative cases and are available upon request.

Figures \ref{polygamy_5} and \ref{polygamy_sentences} display the corresponding results at two granularities: 5-gram sequences, Figure~\ref{polygamy_5},  and sentence-level segments, Figure~\ref{polygamy_sentences}.   In both figures, the colored text (green) highlights the portions previously marked by experts as potentially similar. Moreover, in the heatmap red cells represent pairs with stronger similarity, while yellow
and green cells indicate weaker similarity.  The computed F1-scores at their peak exceed 0.6 for both the 5-gram and sentence-level analyses, indicating a measurable degree of similarity consistent with possible influence.

In Figure \ref{polygamy_5}, one specific 5-gram segment pair ("of a plurality of husbands" versus "a plurality of husbands, instead") produces a notably high F1-score (0.77), reflecting a strong degree of local similarity between the two texts. This elevated score likely results from overlapping lexical tokens shared between the work Melville read in the book by Stewart and the one he later authored. In Figure \ref{polygamy_sentences}, the sentences from these same text portions also display comparatively higher similarity values (0.63), although it was slightly lower than in the n-gram analysis. This difference arises because sentence-level comparisons encompass more tokens, and the similarity tends to be concentrated within particular portions of the sentence rather than uniformly distributed across the entire segment.

\begin{figure*}[h!]
\centering
\includegraphics[width=.9\textwidth]{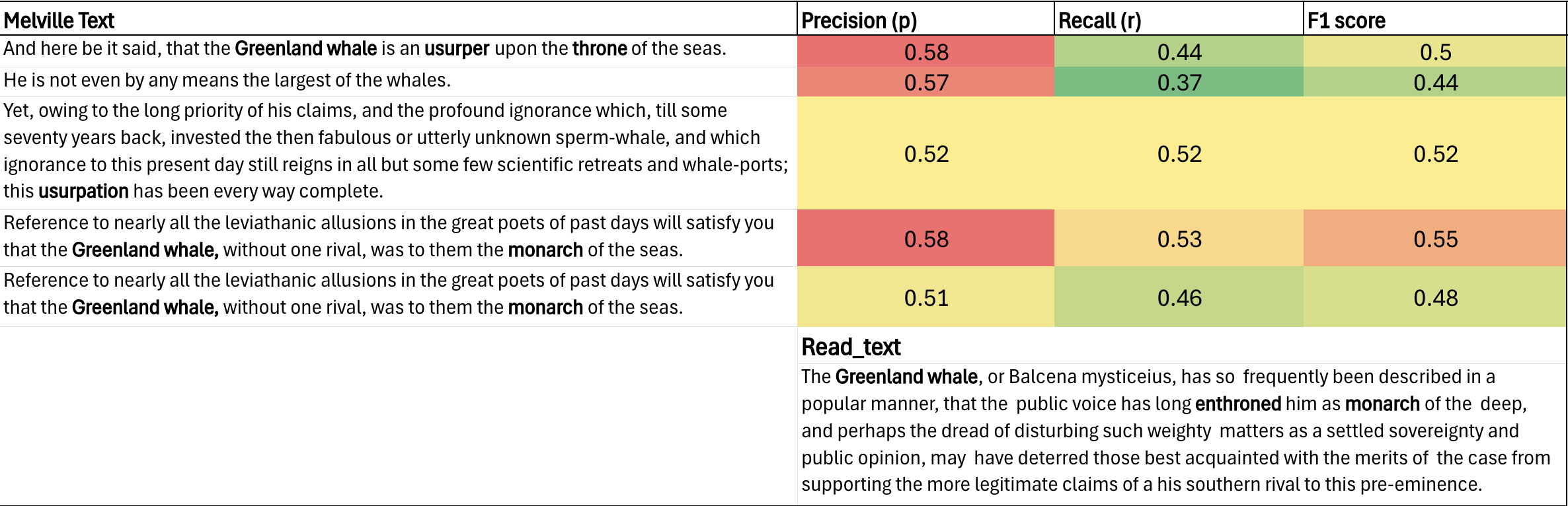}
\caption{Precision, recall and F1 score values for instance 4 from Table \ref{instances} comparing \textit{The Natural History of the Sperm Whale} and \textit{Moby Dick}}
\label{2nd instance fig}
\end{figure*}

The example in Figure \ref{polygamy_sentences} illustrates similarity at the sentence level, where the sentence in Melville’s writing appears to be influenced by a single sentence from the source text. The Table \ref{instances}  instance 1 and instance 2 text pairs are both examples of the sentence to sentence similarity scenario. However, other scenarios occur in which one written sentence is influenced by  several sentences in the source. In Table \ref{instances} instances 3 and 4 represent such a scenario.   Figure~\ref{2nd instance fig}, shows the result from instance 4 of Table \ref{instances} where text from \textit{Moby Dick} is compared with text from \textit{The Natural History of Sperm Whale}. For this example, the values for all metrics (precision, recall, and F1 score) are presented in Figure~\ref{2nd instance fig}. 
The orientation of read vs written text is swapped for better display.
It is important to note that sentence-to-sentence influence generally produces high similarity values, as seen in Figure~\ref{polygamy_sentences}. In contrast, when the same idea is distributed across multiple sentences, the similarity signal is diluted because each individual sentence contains only a portion of the shared meaning, which lowers the sentence-level similarity score as seen in Figure \ref{2nd instance fig}. Similar sentence to multiple sentence influence was identified in instance 3 from Table \ref{instances}, showing the same results of diluted similarity signal, but results are not included in this paper.  The difference in precision, recall and F1 scores are also important here. The higher precision indicates that most tokens in Melville’s sentence matched strongly with those in the source. 
The comparatively lower recall indicates the source contains more content that was not reflected in Melville's sentence, as the idea read in the source sentence was distributed and explained in multiple sentences in Melville's writing. The example strongly illustrates the importance of subjecting computational results of this procedure to human expert examination. As an utterance inspired by multiple successive sentences in the work Melville read, the passage is deeply indebted to the source. It's rhetoric, moreover, illustrates Melville's frequent tendency in source appropriation to consciously "amplify" or "augment" stylistically the attitude or claim conveyed in the original text--a practice that parallels his related method of reversing or opposing the attitude of a source, as seen in the example illustrated  by Figure \ref{polygamy_5} \cite{olsen2011melville, norberg2023technical}. 
Yet the low values in the output of Figure \ref{2nd instance fig} do not on their face indicate a strong correlation. 

\subsection{Model-Discovered Parallels: Additional Evidence of Influence}
\begin{figure*}[h!]
\centering
\includegraphics[width=0.8\textwidth]{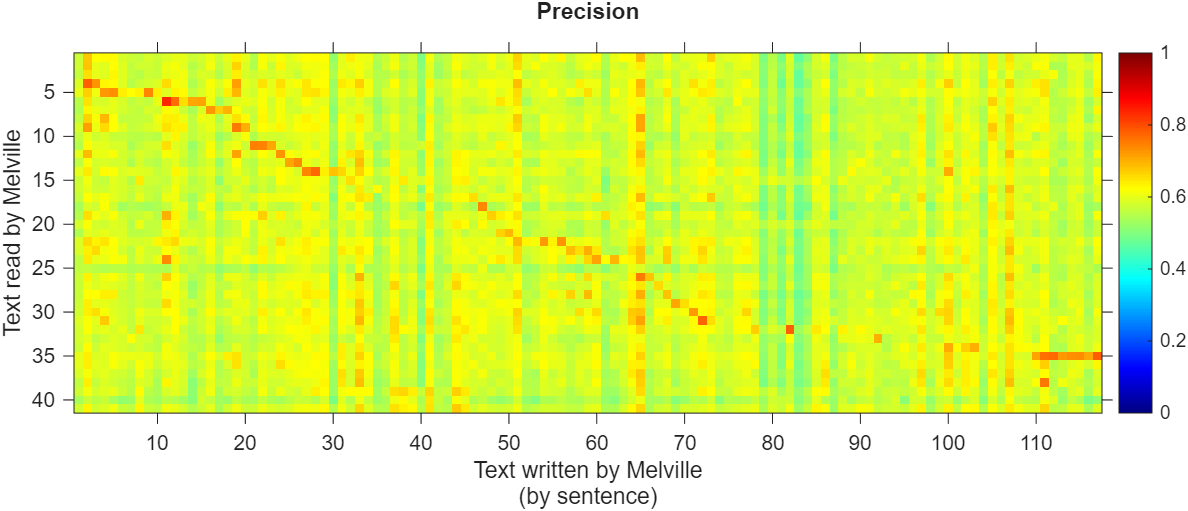}
\caption{Heat map for complete text comparison of instance 2 in  Table \ref{instances} comparing \textit{Life and Remarkable Adventures of Israel R. Potter} with \textit{Israel Potter; His fifty years of exile} at the sentence level.}
\label{full output}
\end{figure*}
After examining the expert-identified cases, we next analyzed the model’s broader output to explore whether it could independently capture additional instances of potential influence or semantic similarity. To this end, we reviewed the complete set of results for text from instance 2, Table \ref{instances} and  present them in Figure \ref{full output} showing a heatmap of precision values  where Melville’s \textit{Israel Potter; His Fifty Years of Exile} was compared with Trumbull’s \textit{Life and Remarkable Adventures of Israel R. Potter}  at the sentence level. The figure illustrates a diverse range of color intensities. As in the previous figures, darker and  red cells represent pairs with stronger similarity, while yellow and green cells indicate weaker similarity.

A few features are visible in this figure. The presence of the high value (red) diagonal line comes from Melville literally retelling the story from the source text for a novel conceived and openly acknowledged by Melville to be a rewriting of Trumbull's 1824 Life \cite{Leyda1951}. In the run of early chapters represented by the passages in Figure 4, Melville is believed to have transcribed words and expressions from Trumbull into approximately half the lines of his lost manuscript, occasionally adding "brief speculations on Israel's motives, or his recurrent foreshadowings of coming tribulations" \cite{bezanson1982historicalnote}. As composition proceeded beyond the early chapters of Israel Potter, Melville allowed himself greater freedom from his source and introduced a variety of imaginative episodes that do not appear in Trumbull's narrative.

To gain deeper insight into these results, we identified the cells corresponding to higher precision values and analyzed the associated text segments.  For instance, from the full output, the highest value in the heatmap occurs at source sentence 6 and Melville's sentence 11 where the precision value reached 0.8. These sentences along with the results are shown as the first text pair example in Table \ref{tab:text_pairs}.   
The BERTScore of 0.80 indicates that most tokens in Melville’s sentence matched strongly with those in the source. However, the recall was lower (0.43) due to the source containing significantly more content that was not reflected in Melville’s sentence. The resulting F1 score was 0.56, suggesting semantic overlap and possible conceptual influence. The second text pair in Table \ref{tab:text_pairs} is taken from the n-gram of 5 experiments from the same text pair instance i.e. instance 2 from Table \ref{instances} (Israel Potter).  Here even at the 5-gram level of granularity the p, r and F1 values are high and the semantic similarity is also evident from the tokens as they reflect similar overall semantic meaning. 

Also visible are some vertical streaks at sentences 65 and 107 in Melville's \textit{Israel Potter; His Fifty Years of Exile}. This highlights one of the limitations: small or generic text tends to yield higher values, which may result in false positives. For example, sentence 65; “About noon the knight visited his workmen” produces relatively high similarity values because it is short and semantically generic, containing broad concepts such as time, authority, and labor. These elements occur frequently across many contexts, allowing embedding-based metrics to find plausible matches even when no specific relationship exists. As a result, this sentence has a higher potential to produce false positives. 
 Similarly sentence 107 \textit{“Now the superintendent of the garden was a harsh, overbearing man.” } yielded relatively high similarity values throughout because it describes a generic authority figure and evaluative traits that commonly appear in narrative texts.

\begin{table*}[!h]
\centering
\caption{Two text pair examples for model discovered parallels from Melville's \textit{Israel Potter; His Fifty Years of Exile} and \textit{Life and Remarkable Adventures of Israel R. Potter}. The first is at the sentence level and the second at the 5-gram level.}
\begin{tabular}{|p{4cm}|p{8.5cm}|p{1.7cm}|}
\hline
\textbf{Melville Text} & \textbf{Read Text} & \textbf{Results} \\
\hline
Israel had now been three days without food, except one two-penny loaf. (sentence 11)&
I had now been three days without food (with the exception of a single two-penny loaf) and felt myself unable much longer to resist the cravings of nature—my spirits, which until now had armed me with fortitude began to forsake me—indeed I was at this moment on the eve of despair! When, calling to mind that grief would only aggravate my calamity, I endeavoured to arm my soul with patience; and habituate myself as well as I could, to woe. Accordingly I roused my spirits; and banishing for a few moments these gloomy ideas, I began to reflect seriously on the methods by which to extricate myself from this labyrinth of horror. (sentence 6) &p~=~0.80 r~=~0.43 F1~=~0.56\\
\hline
presenting a smaller supply than (chunk 301)& which containing a small quantity (chunk 88) &p~=~0.65 r~=~0.70 F1~=~0.68\\
\hline
\end{tabular}
\label{tab:text_pairs}
\end{table*}


\subsection{Evaluating Similarity at Different Levels of Text Granularity}
We have conducted experiments at both the n-gram of 5 and sentence levels to compare how the granularity of text segmentation affects the model’s ability to detect semantic similarity. Both approaches successfully identified instances of correspondence between Melville’s writings and the source texts. The 5-gram analysis provided a more fine-grained view of similarity patterns. As illustrated in Figure~\ref{polygamy_5} and further supported by Table~\ref{tab:text_pairs}. The 5-gram method was particularly effective in capturing localized lexical and semantic overlaps, short phrases or recurring expressions that might indicate influence at the phrase level.

In contrast, the sentence-level analysis yielded a broader contextual comparison but introduced variability in precision and recall values. This variation primarily stems from differences in sentence length and the number of tokens being compared. When one sentence contains significantly more tokens than the other, precision tends to be higher for shorter sentences, while recall decreases because a smaller portion of the longer sentence aligns semantically. The 5-gram method, by comparison, mitigates this issue by keeping token counts relatively consistent across comparisons, allowing for a more balanced representation of local similarity.

Overall, these results suggest that the 5-gram segmentation approach is better suited for detecting localized lexical or phrasal similarities, while the sentence-level analysis provides complementary insights into broader semantic relationships. Using both levels of analysis together would offer a more comprehensive understanding of textual influence, capturing both micro-level echoes and macro-level thematic parallels.




\section{Limitations and Future Work}
This study operates under a few practical constraints that shape the scope of the results. 
First, we used non overlapping five grams that offers an efficient segmentation strategy, although it may overlook similarities that cross unit boundaries. Using overlapping five grams would yield more complete coverage of the text, but doing so substantially increases computational cost. This reflects an unavoidable tradeoff between granularity and computational efficiency.

Second, a further limitation arose from the behavior of BERTScore when comparing very short and generic sentences yielding high precision values because their broad semantic content can be matched to many different contexts in the reference text, even when no meaningful similarity exists. This can also be seen in Figure \ref{full output}  and discussed in detail in Section 4.2. Future work may address this by applying weighting scores by sentence specificity, or incorporating additional models that better distinguish generic similarity from substantive semantic alignment. By selecting a complementary text comparison technique, many false positives could be removed from consideration.

Finally, the analysis was conducted at only two segmentation levels: 5-gram and sentence-level. These levels work well when influence occurs at comparable spans of text, such as the sentence-to-sentence as shown in Figure \ref{polygamy_sentences}. However, when the influence involves mismatched chunks such as a single sentence that draws on multiple sentences or an entire paragraph, sentence-level comparisons may yield lower similarity values simply because the comparison window is not aligned with the underlying structure of the influence. Additional segmentation strategies, including phrase-level and paragraph-level segmentation, may therefore better capture both localized and broader contextual similarities.
\section{Conclusion}
This study explored whether semantic similarity methods, specifically BERTScore, can help identify potential influence between Herman Melville’s writings and the books he read. Using expert-identified examples as a starting point, the analysis showed that BERTScore was able to detect these known similarities and also highlighted additional passages with notable semantic alignment. These findings suggest that semantic similarity measures can provide meaningful insights into possible connections between an author’s reading and writing.


\section*{Acknowledgements} 
This work was supported by the Wallenberg AI, Autonomous Systems and Software Program (WASP), funded by Knut and Alice Wallenberg Foundations and counterpart funding from Luleå University of Technology (LTU).

\bibliography{custom}
\bibliographystyle{acl_natbib}

\appendix

\label{sec:appendix}



\end{document}